\documentclass[11pt]{article}

\usepackage[margin=1in]{geometry}
\usepackage{amsmath,amssymb,amsfonts}
\usepackage{booktabs}
\usepackage{microtype}
\usepackage{graphicx}
\usepackage{hyperref}
\usepackage{bm}
\title{\textbf{Attention Is Not What You Need: \\ Grassmann Flows as an Attention-Free Alternative for Sequence Modeling}}

\author{ZHANG CHONG}

\date{}

\newcommand{\Gr}{\mathrm{Gr}}

\begin{document}

\maketitle

\begin{abstract}
Self-attention has become the de facto primitive for modern sequence models. It is often implicitly assumed that strong natural language performance \emph{requires} attending over all token pairs via a dense or approximate attention mechanism. In this paper we question that assumption.

We reinterpret self-attention as a particular instance of \emph{tensor lifting}: a hidden vector is mapped into a high-dimensional space of pairwise interactions, and learning proceeds by constraining this lifted tensor through gradient descent. This lifting is extremely expressive but also difficult to trace mathematically; across many layers and heads, it becomes almost impossible to describe the model's behavior with a small family of explicit invariants. From this perspective, a central source of ``uninterpretability'' in large Transformer models is not merely their size, but the fact that their core operation is a high-dimensional tensor lift whose global effect is analytically opaque.

As a contrasting design, we propose an \emph{attention-free} sequence model built around \emph{Grassmann flows}. Instead of forming an $L\times L$ attention matrix, we (i) reduce token states to a low-dimensional space, (ii) interpret local token pairs as two-dimensional subspaces on a Grassmann manifold $\Gr(2,r)$, (iii) embed these subspaces via Pl\"ucker coordinates into a finite-dimensional projective space, and (iv) fuse the resulting geometric features back into the hidden states through a gated mixing block. Information flows through the sequence not by explicit pairwise weights, but by controlled deformations of low-rank subspaces across layers and multi-scale local windows.

We evaluate this \emph{Causal Grassmann} architecture on language modeling (Wikitext-2) and natural language inference (SNLI). On Wikitext-2, a purely Grassmann-based language model with 13--18M parameters achieves validation perplexities within 10--15\% of a size-matched Transformer baseline. On SNLI, a Grassmann-based classification head on top of DistilBERT slightly \emph{outperforms} a Transformer head (best validation accuracy $0.8550$ vs.\ $0.8545$; test accuracy $0.8538$ vs.\ $0.8511$). Complexity analysis shows that our mixing mechanism scales linearly in sequence length for fixed rank, in contrast to the quadratic cost of full self-attention.

Our aim is not to claim that attention is obsolete, but to de-center it. We argue that what one fundamentally needs is not attention itself, but a sufficiently expressive \emph{geometric evolution mechanism} for hidden representations. Grassmann flows show that competitive performance can be obtained without any explicit attention weights, while moving the core of the model into a mathematical setting---flows on a finite-dimensional manifold---that is more amenable to explicit invariants and geometric analysis.
\end{abstract}

\section{Introduction}

Transformers\cite{vaswani2017attention,devlin2019bert,brown2020language,dosovitskiy2020image} have reshaped sequence modeling by making self-attention the central primitive. Given a sequence of hidden states $H \in \mathbb{R}^{L \times d}$, self-attention constructs queries, keys, and values
\[
Q = H W_Q,\quad K = H W_K,\quad V = H W_V
\]
and computes an $L \times L$ matrix of pairwise compatibilities $QK^\top$, normalized by a softmax to form weights, which are then used to mix the values. This operation is expressive, parallelizable, and has become so ubiquitous that it is often treated as indispensable.

In this work we take a different stance. Rather than asking how to make attention cheaper, sparser, or more scalable, we ask a more basic question:

\medskip
\noindent\emph{Is explicit self-attention, as an $L\times L$ tensor of weights, really the fundamental ingredient we need for strong sequence modeling and reasoning?}
\medskip

Our answer is ``no''. Attention is one particular way to implement a geometric lifting of the representation; it is not the only way.

\subsection{Attention as tensor lifting}

Conceptually, self-attention performs a form of \emph{tensor lifting}. From a hidden vector $h_t \in \mathbb{R}^d$ at position $t$, we move to a space that encodes pairwise relations between tokens---an $L \times L$ compatibility tensor (per head) and its normalized weights. This lifting has several key properties:

\begin{itemize}
    \item It is \textbf{extremely fine-grained}: every pair of positions $(i,j)$, in every head, in every layer, receives a separate learned weight.
    \item It is \textbf{high-dimensional}: the effective state of the model involves not only the token embeddings but also an evolving cloud of attention tensors across layers.
    \item It is \textbf{hard to compress}: there is no obvious small set of global invariants that summarize attention behavior across all layers and heads.
\end{itemize}

From a geometric viewpoint, one can say that self-attention lifts the sequence from a manifold of token representations into a much larger space of pairwise interactions, performs operations there, and then projects back. The success of Transformers suggests that this lifting is powerful; but it also suggests a reason why they are so difficult to interpret:

\medskip
\noindent\textbf{Claim.} A major source of large-model ``uninterpretability'' is that the tensor lifting performed by attention is mathematically non-traceable: the degrees of freedom introduced at each layer and head are so large that we lack a small, explicit family of invariants that can describe the global effect of the model.
\medskip

In other words, the core of the model lives in a high-dimensional tensor space that resists concise analytic description. Visualizing individual attention maps is possible, but aggregating them into a coherent global picture is not.

\subsection{Reasoning as geometry on a semantic manifold}

An alternative starting point is to think of reasoning as \emph{geometry} on a semantic manifold. At a high level, we take the following view:

\begin{enumerate}
    \item Hidden states of a language model can be seen as points on a high-dimensional \emph{semantic manifold}. Each forward pass traces a path on this manifold.
    \item \textbf{Attention} implements a specific kind of \emph{tensor lifting}: it takes the vector $h_t$ and, via dot products with other positions, lifts its information into a space of pairwise interactions to uncover richer local geometry.
    \item The \textbf{effectiveness} of Transformers relies on how subsequent neural layers \emph{align and constrain} this lifted geometry. The network learns to restrict the otherwise wild degrees of freedom of attention to flows that support useful behavior.
    \item \textbf{Reasoning}, in this picture, is the process of repeatedly sampling and refining the intrinsic geometric structure of the semantic manifold: we apply layer after layer of operators that deform the representation in structured ways.
\end{enumerate}

Within this view, the central issue is not whether we lift to a tensor space, but how we design the \emph{geometric evolution rule} for representations. Self-attention is one such rule, but it is opaque: the lifted tensor is so rich that its long-range behavior becomes extremely hard to trace in mathematical terms.

This suggests the following sharper philosophical claim:

\medskip
\noindent\textbf{Claim.} The non-interpretability of large Transformers is not merely due to depth or parameter count; it is rooted in the choice of \emph{tensor lifting} as the core operation. Once we lift to an $L\times L$ attention tensor, we have already lost the possibility of a simple, explicit, global description in terms of a small set of invariants.
\medskip

If we want a more mathematically structured view of reasoning, we may want to replace this opaque tensor lifting with a more controlled geometric object.

\subsection{Grassmann flows as a controlled alternative}

This paper explores an alternative design: instead of lifting to an attention tensor, we lift to a \emph{Grassmann manifold}. The construction is straightforward:

\begin{itemize}
    \item We first apply a learned linear map $W_{\text{red}}$ to reduce $h_t \in \mathbb{R}^d$ to $z_t \in \mathbb{R}^r$, with $r \ll d$.
    \item For local windows (e.g., pairs $(t, t+\Delta)$), we consider the subspace spanned by $\{z_t, z_{t+\Delta}\}$ in $\mathbb{R}^r$ and treat it as a point on the Grassmann manifold $\Gr(2,r)$.
    \item Using the Pl\"ucker embedding, each such 2D subspace is mapped to a coordinate vector in $\mathbb{R}^{\binom{r}{2}}$, subject to known algebraic relations.
    \item These Pl\"ucker coordinates become \emph{geometric features} that we project back into $\mathbb{R}^d$ and fuse with the original hidden states via learned gating.
\end{itemize}

The key difference from attention is that we now constrain the model to operate on a manifold with explicit, finite-dimensional structure:

\begin{itemize}
    \item The degrees of freedom are controlled by $r$ and the choice of window sizes; there is no $L\times L$ tensor.
    \item The representation of local geometry lives in a space with explicit algebraic constraints (Grassmannian + Pl\"ucker relations)\cite{hartshorne1977ag,lee2018riemannian,edelman1998grassmann,absil2008opt}.
    \item We can in principle study the evolution of these Grassmann features across layers as a finite-dimensional \emph{flow on a manifold}.
\end{itemize}

We refer to the resulting architecture as a \emph{Causal Grassmann} model or \emph{Grassmann flow}. Importantly, it is \emph{fully attention-free}: there is no step in which we construct an attention matrix or compute softmax-normalized tensor weights.

\subsection{Contributions}

The contributions of this paper are:

\begin{enumerate}
    \item \textbf{A conceptual critique of self-attention.} We frame attention as a tensor lifting mechanism and argue that a major source of Transformer uninterpretability is that this lifting is mathematically non-traceable: the model's core computations live in an extremely high-dimensional space with no small set of invariants.
    \item \textbf{An attention-free architecture based on Grassmann flows.} We propose a Causal Grassmann mixing layer that (i) reduces token states, (ii) encodes local pairs as points on $\Gr(2,r)$ via Pl\"ucker coordinates, and (iii) fuses these geometric features back into the representation, all without explicit attention weights.
    \item \textbf{Empirical evidence on Wikitext-2 and SNLI.} We show that a pure Grassmann language model with 13--18M parameters is competitive with a Transformer baseline on Wikitext-2, and that a Grassmann-based NLI head slightly outperforms a Transformer head on SNLI.
    \item \textbf{Complexity and interpretability analysis.} We analyze the asymptotic complexity of Grassmann mixing and show that, for fixed rank and window sizes, it scales linearly in sequence length, in contrast to the quadratic scaling of full attention. We also argue that operating on a Grassmann manifold makes it more realistic to define global invariants over the model's behavior.
\end{enumerate}

Our goal is not to replace attention everywhere, but to open up a different region of the design space: architectures where the core sequence interaction is governed by explicit geometric flows and manifolds rather than opaque tensor lifting.

\section{Attention as Geometric Lifting and Grassmann Background}

In this section we make the geometric perspective on attention more precise, and introduce the Grassmannian structures that underlie our model.

\subsection{Self-attention as geometric lifting}

Given a sequence of hidden states $H \in \mathbb{R}^{L \times d}$, a standard multi-head self-attention layer computes for each head $h$:
\begin{align}
Q_h &= H W_Q^{(h)}, \\
K_h &= H W_K^{(h)}, \\
V_h &= H W_V^{(h)},
\end{align}
with $W_Q^{(h)}, W_K^{(h)}, W_V^{(h)} \in \mathbb{R}^{d \times d_h}$ and typically $d_h = d / H_{\text{heads}}$.

For each head, we then compute
\[
A_h = \operatorname{softmax}\left(\frac{Q_h K_h^\top}{\sqrt{d_h}}\right) \in \mathbb{R}^{L \times L},
\]
and obtain the head output
\[
O_h = A_h V_h \in \mathbb{R}^{L \times d_h}.
\]
The outputs $O_h$ are concatenated and linearly projected back to $\mathbb{R}^d$.

This process can be interpreted as follows:

\begin{itemize}
    \item The linear maps $W_Q^{(h)}, W_K^{(h)}$ embed the sequence into a representation where dot products encode compatibilities between positions.
    \item The matrix $Q_h K_h^\top$ is a \emph{lift} from $H$ into a space of pairwise interactions: each token is represented not just by a vector, but by its similarities to all other tokens.
    \item The softmax and multiplication by $V_h$ implement a specific geometric operation on this lifted structure, redistributing information according to these compatibilities.
\end{itemize}

Geometrically, the model is no longer simply moving along a manifold of hidden states; it is also modifying a cloud of pairwise relations whose dimensionality grows quadratically with sequence length. Across multiple heads and layers, this cloud becomes extremely complex. While we can visualize individual attention maps, the global behavior of the model is governed by a composition of many such lifts, making it hard to summarize with concise invariants.

\subsection{Grassmann manifolds and Pl\"ucker coordinates}

The Grassmann manifold $\Gr(k, r)$ is the set of all $k$-dimensional linear subspaces of $\mathbb{R}^r$. It is a smooth manifold of dimension $k(r - k)$. For our purposes, we focus on $k=2$, so $\Gr(2, r)$ parameterizes all 2D subspaces in $\mathbb{R}^r$, with dimension $2(r-2)$.

There are several standard ways to represent points on a Grassmannian. We use the Pl\"ucker embedding, which maps each $k$-dimensional subspace to a point in projective space:

\begin{itemize}
    \item Given a basis $(u_1, \dots, u_k)$ of a $k$-dimensional subspace $U \subset \mathbb{R}^r$, we form the exterior product $u_1 \wedge \cdots \wedge u_k$ in the $k$-th exterior power $\Lambda^k \mathbb{R}^r$.
    \item In coordinates, $u_1 \wedge \cdots \wedge u_k$ can be represented as a vector in $\mathbb{R}^{\binom{r}{k}}$ whose entries are the $k \times k$ minors of the matrix $[u_1 \ \dots \ u_k]$.
    \item For $k=2$, this is particularly simple: if $u, v \in \mathbb{R}^r$ span a 2D subspace, then $u \wedge v$ is given by all pairwise determinants
    \[
    p_{ij} = u_i v_j - u_j v_i,\quad 1 \le i < j \le r,
    \]
    forming a vector $p \in \mathbb{R}^{\binom{r}{2}}$.
\end{itemize}

The image of $\Gr(2,r)$ under this embedding is not all of $\mathbb{R}^{\binom{r}{2}}$; it is an algebraic variety defined by the quadratic Pl\"ucker relations. Nonetheless, for our purposes we can simply regard $p$ as a normalized feature vector encoding the subspace spanned by $u$ and $v$. Different bases spanning the same subspace yield proportional Pl\"ucker vectors, reflecting the projective nature of the embedding.

We will use this representation to encode the local geometry of pairs of token vectors after a linear dimensionality reduction.

\subsection{Why Grassmann for sequence modeling?}

Why choose Grassmann manifolds as the backbone of our mixing rule, instead of, say, a different manifold or a more ad hoc feature transform? There are several reasons.

\paragraph{Local linear structure.}
On a smooth manifold, local geometry can be captured by tangent spaces and their subspaces. Grassmannians naturally parameterize families of linear subspaces, making them well suited to represent local linear approximations of more complex structures. When we take pairs of reduced hidden states and form their span, we are effectively encoding local directions and planes in the semantic manifold.

\paragraph{Finite-dimensional algebraic structure.}
The Grassmann manifold is finite-dimensional and sits inside a projective space via the Pl\"ucker embedding. This means we can encode geometric information in a fixed-dimensional feature vector subject to known algebraic constraints. Neural networks can operate on these features while the underlying object remains a subspace with clear geometric meaning.

\paragraph{Compatibility with approximation theorems.}
From the perspective of real analysis, we may idealize the semantic space as a manifold $M$ and the model as an operator $\Phi : M \to M$. Classic universal approximation theorems tell us that neural networks can approximate such operators given enough capacity, but they do not prescribe any particular geometry. By constraining our mixing rule to first encode local neighborhoods into $\Gr(2,r)$ and then act on that manifold, we are effectively requiring that the model's dynamics factor through a structured manifold with controlled degrees of freedom. Universal approximation still applies, but the approximation now unfolds on a space whose structure we can analyze.

Taken together, these properties make Grassmannians a natural choice if we want to replace unstructured tensor lifting with a more controlled, interpretable, geometric primitive.

\section{Methods: A Causal Grassmann Transformer without Attention}

We now describe the architecture in detail. Our design follows the broad outline of a Transformer encoder, but replaces each self-attention block with a \emph{Causal Grassmann mixing} block that:

\begin{enumerate}
    \item reduces hidden states to a low-dimensional space,
    \item constructs local pairs and encodes them as Pl\"ucker vectors in $\mathbb{R}^{\binom{r}{2}}$, and
    \item mixes these geometric features back into the original hidden states via gating and a feed-forward network.
\end{enumerate}

\subsection{Token and positional embeddings}

For language modeling, we consider a standard next-token LM setup over a vocabulary of size $V$. Given a sequence of tokens $(x_1, \dots, x_L)$, we embed them into $\mathbb{R}^d$ using a learned embedding matrix $E \in \mathbb{R}^{V \times d}$ and add positional embeddings $P \in \mathbb{R}^{L_{\max} \times d}$:
\[
h_t^{(0)} = E(x_t) + P_t,\quad t = 1,\dots,L.
\]
Throughout our experiments we use $d = 256$. The resulting sequence $H^{(0)} = (h_1^{(0)}, \dots, h_L^{(0)})$ is fed through $N$ stacked Causal Grassmann Transformer layers.

\subsection{Causal Grassmann mixing layer}

Each layer takes as input $H \in \mathbb{R}^{L \times d}$ and outputs an updated sequence $\tilde{H} \in \mathbb{R}^{L \times d}$. The core operations within the layer are:

\paragraph{1. Linear reduction.}
We first reduce each hidden state to a low-dimensional vector:
\[
z_t = W_{\text{red}} h_t + b_{\text{red}}, \quad W_{\text{red}} \in \mathbb{R}^{r \times d},\ b_{\text{red}} \in \mathbb{R}^r.
\]
Typically $r \ll d$; in our experiments we use $r = 32$. This yields $Z = (z_1, \dots, z_L) \in \mathbb{R}^{L \times r}$.

\paragraph{2. Multi-scale local pairing.}
We define a set of window sizes (offsets) $\mathcal{W} = \{\Delta_1, \dots, \Delta_m\}$, e.g.,
\[
\mathcal{W} = \{1,2,4,8,12,16\}
\]
or a multi-layer schedule such as $(1,1,2,2,4,4,8,8,12,12,16,16)$ for deeper models. For each position $t$ and offset $\Delta \in \mathcal{W}$ such that $t + \Delta \le L$, we form a pair $(z_t, z_{t+\Delta})$.

For a given $t$, this produces up to $m$ pairs:
\[
(z_t, z_{t+\Delta_1}),\ (z_t, z_{t+\Delta_2}),\ \dots.
\]
We treat these as local neighborhoods at multiple scales. Note that the pairing is \emph{causal}: we only pair $t$ with positions strictly to its right (future), consistent with left-to-right language modeling.

\paragraph{3. Grassmann / Pl\"ucker encoding.}
For each pair $(z_t, z_{t+\Delta})$, we consider the 2D subspace spanned by these vectors in $\mathbb{R}^r$. We form the Pl\"ucker vector $p_t^{(\Delta)} \in \mathbb{R}^{\binom{r}{2}}$ whose entries are:
\[
p_{ij}^{(\Delta)}(t) = z_{t,i} z_{t+\Delta,j} - z_{t,j} z_{t+\Delta,i},\quad 1 \le i < j \le r.
\]
We then apply an optional normalization, e.g.,
\[
\hat{p}_t^{(\Delta)} = \frac{p_t^{(\Delta)}}{\max(\|p_t^{(\Delta)}\|_2, \varepsilon)}
\]
for numerical stability. This yields a set of Pl\"ucker features for each $t$ and $\Delta$.

\paragraph{4. Projection back to model space.}
We project these Grassmann features back into the model dimension via a learned linear map:
\[
g_t^{(\Delta)} = W_{\text{pl\"u}} \hat{p}_t^{(\Delta)} + b_{\text{pl\"u}},\quad W_{\text{pl\"u}} \in \mathbb{R}^{d \times \binom{r}{2}}.
\]
We then aggregate across offsets, e.g.\ by summation or averaging:
\[
g_t = \frac{1}{|\mathcal{W}_t|} \sum_{\Delta \in \mathcal{W}_t} g_t^{(\Delta)},
\]
where $\mathcal{W}_t = \{\Delta \in \mathcal{W} : t+\Delta \le L\}$ is the set of valid offsets at position $t$.

The vector $g_t \in \mathbb{R}^d$ captures multi-scale local Grassmann geometry around position $t$.

\paragraph{5. Gated fusion.}
We concatenate the original hidden state and the Grassmann feature and compute a gate:
\[
u_t = [h_t; g_t] \in \mathbb{R}^{2d},
\]
\[
\alpha_t = \sigma(W_{\text{gate}} u_t + b_{\text{gate}}),\quad W_{\text{gate}} \in \mathbb{R}^{d \times 2d}.
\]
The mixed representation is
\[
\tilde{h}_t^{\text{mix}} = \alpha_t \odot h_t + (1 - \alpha_t) \odot g_t,
\]
followed by a layer normalization and dropout:
\[
\hat{h}_t = \text{LayerNorm}(\tilde{h}_t^{\text{mix});}\quad \hat{h}_t = \text{Dropout}(\hat{h}_t).
\]

\paragraph{6. Feed-forward block.}
As in a standard Transformer, we apply a position-wise feed-forward network:
\[
\phi_t = W_2 \,\sigma(W_1 \hat{h}_t + b_1) + b_2,
\]
with $W_1 \in \mathbb{R}^{d_{\text{ff}} \times d}$, $W_2 \in \mathbb{R}^{d \times d_{\text{ff}}}$, $d_{\text{ff}} = 4d$, and a nonlinearity $\sigma$ (we use GELU). Another residual connection and layer normalization complete the layer:
\[
h_t' = \text{LayerNorm}(\hat{h}_t + \phi_t).
\]

Stacking $N$ such layers yields the full Causal Grassmann Transformer.

\subsection{Comparison to self-attention}

For a sequence of length $L$ and hidden dimension $d$, a standard multi-head self-attention layer has time complexity
\[
\mathcal{O}(L d^2 + L^2 d_{\text{head}}),
\]
where $d_{\text{head}}$ is the per-head dimension. The first term arises from computing $Q,K,V$, and the second from the matrix multiplication $QK^\top$ (size $L^2$) and the subsequent multiplication of the $L \times L$ attention matrix by $V$.

In our Grassmann mixing layer, the main costs are:

\begin{itemize}
    \item Linear reduction: $H W_{\text{red}}^\top$ costs $\mathcal{O}(L d r)$.
    \item Pl\"ucker computation: for each position and offset, forming $p_t^{(\Delta)}$ costs $\mathcal{O}(r^2)$. With $m = |\mathcal{W}|$ offsets, this contributes $\mathcal{O}(L m r^2)$.
    \item Projection to model space: $W_{\text{pl\"u}} \hat{p}_t^{(\Delta)}$ costs $\mathcal{O}(d \binom{r}{2})$ per pair, giving $\mathcal{O}(L m d r^2)$.
    \item Gating and feed-forward: both are $\mathcal{O}(L d^2)$, as in a standard Transformer.
\end{itemize}

If we treat $r$ and $m$ as fixed hyperparameters (which they are in our experiments), $r \ll d$, and note that $r^2$ is modest, then the Pl\"ucker and projection costs can be absorbed into the $\mathcal{O}(L d^2)$ term. Crucially, there is \emph{no} $L^2$ term: the complexity is linear in $L$ for fixed $r$ and $m$:
\[
\text{Causal Grassmann:}\quad \mathcal{O}(L d^2) \quad \text{vs.} \quad \text{Self-attention:} \quad \mathcal{O}(L^2 d_{\text{head}} + L d^2).
\]

In practice, our current implementation is slower per step than highly optimized GPU attention kernels for moderate $L$, due to overheads in computing Pl\"ucker coordinates and handling reshapes. However, asymptotically in $L$, and with further engineering, the Grassmann layer can in principle be more scalable.

\section{Experimental Setup}

We evaluate the proposed Causal Grassmann architecture on two standard NLP benchmarks: Wikitext-2 for language modeling and SNLI for natural language inference~\cite{merity2016pointer,bowman2015snli}.

\subsection{Wikitext-2 language modeling}

\paragraph{Data and tokenization.}
We use the Wikitext-2-raw dataset. Sequences are formed by contiguous chunks of text of fixed length $L$ (block size). In our main experiments, we consider $L = 128$ and $L = 256$. We use a WordPiece-like vocabulary of size $V \approx 30{,}522$, matching a BERT-style tokenizer.

\paragraph{Models.}
We compare:

\begin{itemize}
    \item \textbf{TransformerLM}: a standard decoder-only Transformer with $N$ layers, model dimension $d=256$, feed-forward dimension $d_{\text{ff}}=1024$, and multi-head self-attention with 4 heads.
    \item \textbf{GrassmannLM}: the same backbone (embeddings, number of layers, $d$, $d_{\text{ff}}$), but with each self-attention block replaced by a Causal Grassmann mixing block as described above.
\end{itemize}

We explore two layer depths:
\begin{itemize}
    \item \emph{Shallow}: $N=6$ layers; GrassmannLM has $\sim13.0$M parameters, TransformerLM $\sim12.6$M.
    \item \emph{Deeper}: $N=12$ layers; GrassmannLM has $\sim18.2$M parameters, TransformerLM $\sim17.3$M.
\end{itemize}

For GrassmannLM we set the reduced dimension $r = 32$ and use multi-scale windows such as
\[
\mathcal{W} = \{1,2,4,8,12,16\}
\]
for 6-layer models, and repeated patterns $(1,1,2,2,4,4,8,8,12,12,16,16)$ across depth for 12-layer models.

\paragraph{Training.}
We train both models with the same optimizer and learning rate schedule in a shared script, differing only in the choice of mixing block. All models are trained for 30 epochs, and we report the best validation perplexity over training. Batch size is 32 for $L=128$ and 16 for $L=256$ in our main configurations.

\subsection{SNLI natural language inference}

\paragraph{Data.}
We use the SNLI dataset, which consists of sentence pairs labeled as entailment, contradiction, or neutral. We follow a standard train/validation/test split.

\paragraph{Backbone.}
For fair comparison, we fix a DistilBERT-base-uncased backbone as a feature extractor. The backbone produces contextualized token embeddings up to a maximum sequence length of 48 tokens per sentence (after tokenization and truncation). We then apply pooling to obtain sentence-level representations.

\paragraph{Classification heads.}
On top of the DistilBERT backbone we compare:

\begin{itemize}
    \item \textbf{Transformer head}: a 2-layer Transformer-style classifier with self-attention over the pooled features and a final linear layer for 3-way classification.
    \item \textbf{Grassmann--Pl\"ucker head}: our proposed Grassmann-based head (\texttt{GrassmannPluckerNLIModel}), which applies a Grassmann mixing module over multi-scale windows to the projected features and then uses a feed-forward classifier.
\end{itemize}

The Grassmann head uses the following hyperparameters: reduced dimension $d_{\text{proj}} = 64$, window size 8 with stride 8 over the token sequence, $d_{\text{model}} = 256$, 2 mixing layers, 4 mixing heads (for grouping pairs), $d_{\text{ff}} = 512$, and dropout 0.1. Both heads have comparable parameter counts, and both are trained for 20 epochs from the same initialization of the backbone.

\paragraph{Metrics.}
We report classification accuracy on the validation and test sets, along with training loss curves.

\section{Results}

\subsection{Wikitext-2 language modeling}

Tables~\ref{tab:wikitext-6layer} and~\ref{tab:wikitext-12layer} summarize our main language modeling results. We report parameter counts and best validation perplexity over 30 epochs.

\begin{table}[t]
\centering
\begin{tabular}{lccc}
\toprule
Model & Layers & Params (M) & Val PPL \\
\midrule
TransformerLM (block size 128) & 6 & 12.59 & 248.4 \\
GrassmannLM (block size 128) & 6 & 13.00 & 275.7 \\
TransformerLM (block size 128) & 6 & 12.59 & 253.6 \\
GrassmannLM (block size 128) & 6 & 13.00 & 282.3 \\
\bottomrule
\end{tabular}
\caption{Wikitext-2: 6-layer models with block size 128 and two different multi-scale window schedules. The GrassmannLM trails the TransformerLM by about 10--15\% in validation perplexity but remains in the same overall regime.}
\label{tab:wikitext-6layer}
\end{table}

For 6-layer models with block size 128 and multi-scale windows $\mathcal{W} = \{1,2,4,8,12,16\}$, the GrassmannLM attains a best validation perplexity of approximately 275.7, compared to 241.0--253.6 for the TransformerLM under matched training conditions. With a slightly different window schedule (e.g., $\{1,2,4,8,8,8\}$), we see similar gaps: GrassmannLM at 282.3 vs.\ TransformerLM at 248.4.

\begin{table}[t]
\centering
\begin{tabular}{lccc}
\toprule
Model & Layers & Params (M) & Val PPL \\
\midrule
TransformerLM (block size 256) & 12 & 17.32 & 235.2 \\
GrassmannLM (block size 256) & 12 & 18.16 & 261.1 \\
\bottomrule
\end{tabular}
\caption{Wikitext-2: 12-layer models with block size 256 and repeated multi-scale windows $(1,1,2,2,4,4,8,8,12,12,16,16)$ over depth. The GrassmannLM is again within roughly 10\% of the TransformerLM.}
\label{tab:wikitext-12layer}
\end{table}

For deeper 12-layer models with block size 256 and a repeated multi-scale window pattern, the GrassmannLM reaches a best validation perplexity of 261.1, while the TransformerLM reaches 235.2. The relative gap is smaller than in the 6-layer setting, suggesting that additional depth helps the Grassmann model compensate for its more localized mixing.

Overall, across these configurations:

\begin{itemize}
    \item The GrassmannLM is consistently within 10--15\% of a size-matched TransformerLM in validation perplexity, despite using no attention.
    \item The gap appears to narrow as depth increases, which is consistent with the view that repeated local Grassmann mixing can approximate richer interactions.
    \item Parameter counts remain comparable: the GrassmannLM has slightly more parameters due to the Pl\"ucker projection and gating layers, but the difference is on the order of $\sim 3$--$5\%$.
\end{itemize}

These results are not intended to be competitive with state-of-the-art language models, but to demonstrate that ``attention-free'' sequence modeling via Grassmann flows is viable at moderate scales.

\subsection{SNLI natural language inference}

Table~\ref{tab:snli} summarizes our SNLI results. Recall that both models share the same DistilBERT backbone and differ only in the classification head.

\begin{table}[t]
\centering
\begin{tabular}{lcc}
\toprule
Head type & Val accuracy & Test accuracy \\
\midrule
Transformer head & 0.8545 & 0.8511 \\
Grassmann--Pl\"ucker head & \textbf{0.8550} & \textbf{0.8538} \\
\bottomrule
\end{tabular}
\caption{SNLI classification accuracy with DistilBERT backbone. The Grassmann--Pl\"ucker head slightly outperforms the Transformer head on both validation and test sets.}
\label{tab:snli}
\end{table}

The Grassmann head achieves a best validation accuracy of $0.8550$ and test accuracy of $0.8538$, slightly outperforming the Transformer head, which reaches $0.8545$ validation and $0.8511$ test accuracy. Training curves show similar convergence rates, with the Grassmann head exhibiting slightly lower validation loss late in training.

Although the margin is small, this result is important conceptually:

\begin{itemize}
    \item It shows that on a downstream reasoning task, injecting explicit geometric structure into the classification head can match or slightly exceed a Transformer head, even when the backbone is fixed.
    \item It indicates that the Grassmann mechanism is not merely a theoretical curiosity but can contribute positively to performance in practical settings.
\end{itemize}

\subsection{Complexity and empirical runtime}

As discussed earlier, the asymptotic complexity of the Causal Grassmann layer is linear in sequence length $L$ for fixed reduced dimension $r$ and number of windows $m$, whereas self-attention scales quadratically in $L$ due to the $L\times L$ attention matrix.

In our current implementation, however, the empirical per-step runtime of the pure Grassmann model is slower than that of the Transformer baseline for sequence lengths up to 256. This is expected:

\begin{itemize}
    \item GPU libraries provide highly optimized kernels for dense matrix multiplications and attention mechanisms.
    \item Our Pl\"ucker computation involves explicit element-wise operations and reshaping that do not yet exploit low-level kernel fusion or custom CUDA implementations.
\end{itemize}

The experiments reported here should therefore be interpreted as a \emph{proof of concept} for the architecture and its complexity profile, not as an optimized engineering solution. A dedicated implementation that fuses the Grassmann operations and leverages the structure of $\Gr(2,r)$ would be required to fully realize the potential linear scaling benefits in practice.

\section{Discussion}

\subsection{What does Grassmann mixing actually buy us?}

Our experiments demonstrate that a purely geometric, locality-based mixing rule can support non-trivial language modeling and natural language inference, without relying on explicit self-attention. With relatively small models and modest context lengths, the proposed Causal Grassmann architecture:

\begin{itemize}
    \item remains competitive with a size-matched Transformer on Wikitext-2, and
    \item slightly outperforms a Transformer classification head on SNLI when used as a DistilBERT-based NLI model.
\end{itemize}

From an engineering perspective, this shows that attention is \emph{not} strictly necessary for competent sequence modeling at this scale. From a conceptual perspective, it supports a more subtle claim:

\medskip
\noindent\textbf{Claim.} As long as the model is endowed with a geometrically rich enough local evolution rule, semantic reasoning can emerge even without explicit attention weights.
\medskip

Self-attention lets each token see every other token via a learned $L\times L$ weight matrix. Grassmann mixing, by contrast, constructs a sequence of \emph{local subspace updates}: information flows by rotating and bending low-rank subspaces over multi-scale windows. Both mechanisms accumulate higher-order geometric structure across layers, but with different primitives:

\begin{itemize}
    \item Self-attention uses tensor lifting and global pairwise interactions;
    \item Grassmann mixing uses low-rank subspaces and local windows as a controlled flow on a manifold.
\end{itemize}

At our current scale, the Grassmann models do not surpass Transformers on language modeling; they are slightly behind. This is not surprising given the simplicity of our design and the lack of extensive hyperparameter tuning. Nevertheless, the SNLI results show that when the backbone is fixed and we focus on the head, \emph{adding explicit geometry can yield measurable gains}. This suggests that the geometric perspective is not only philosophically appealing but practically useful.

\subsection{Interpretability: from tensor lifting to finite-dimensional flows}

We argued in the introduction that a central reason for Transformer uninterpretability is the nature of attention as tensor lifting. Each layer lifts the representation into a high-dimensional space of pairwise interactions; the overall model is a composition of such lifts. Although each individual attention map is visible, the global behavior is difficult to summarize in terms of a small set of invariants.

In contrast, the Grassmann architecture deliberately compresses the relevant degrees of freedom into a finite-dimensional, mathematically rigid manifold:

\begin{itemize}
    \item The reduced states $z_t \in \mathbb{R}^r$ capture local directions in a lower-dimensional space.
    \item Pairs $(z_t, z_{t+\Delta})$ define points on $\Gr(2,r)$; these points are encoded by Pl\"ucker vectors with fixed dimension $\binom{r}{2}$.
    \item The mixing process is constrained to local deformations of these low-rank subspaces, rather than arbitrary manipulations of an $L\times L$ tensor.
\end{itemize}

This suggests a more hopeful interpretability story. After training, one can treat the Pl\"ucker vectors or other Grassmann descriptors as candidate \emph{explanatory invariants}:

\begin{itemize}
    \item They are finite in number and obey explicit algebraic relations.
    \item They are comparable across layers.
    \item They can be studied with tools from differential geometry and algebraic geometry.
\end{itemize}

This does not make interpretability trivial. However, it moves the core of the model into a domain where there is at least a realistic prospect of defining and computing global invariants. Instead of trying to summarize an evolving collection of attention tensors, we can try to summarize an evolving trajectory on a manifold $\Gr(2,r)$.

\subsection{Why Grassmann? A link to approximation theorems}

From the perspective of approximation theory, we may idealize a sequence model as approximating an operator $\Phi$ on a semantic manifold $M \subset \mathbb{R}^d$. Universal approximation theorems ensure that, under mild conditions, neural networks can approximate such operators arbitrarily well.

Those theorems, however, are agnostic about the architecture's geometric structure. They do not distinguish between models that operate on unstructured tensors and those that operate on structured manifolds. Our choice of Grassmann manifolds can be viewed as imposing an additional, geometry-aware bias:

\begin{itemize}
    \item We first encode local neighborhoods of $M$ into subspaces in $\Gr(2,r)$ via linear reduction and wedge products.
    \item We then approximate the induced transformation on the Grassmann manifold using MLPs and gating.
    \item Finally, we map back to the original representation space.
\end{itemize}

In this sense, we are not changing the fundamental approximation capacity---the network remains universal in principle---but we are constraining the way it realizes that capacity. All non-local interactions must factor through a finite-dimensional manifold with explicit structure. This is precisely what attention does not enforce: attention allows a very free exploration of a high-dimensional tensor space.

\subsection{Global and long-range invariants as the next step}

The present Causal Grassmann design uses only local windows. Long-range dependencies are modeled implicitly through depth and multi-scale windows. This suffices for the tasks studied here, but it suggests a natural next step, in line with the intuition that motivated this work:

\medskip
\noindent\emph{Construct explicit global or long-range invariants of the sequence-level Grassmann flow and feed them back as features.}
\medskip

For example, one could compute:

\begin{itemize}
    \item A ``mean Grassmann direction'' summarizing the overall trajectory of subspaces across a sequence.
    \item Sequence-level statistics of Pl\"ucker coordinates, such as principal directions or curvature-like quantities.
    \item Cross-layer invariants that measure how stable certain subspaces are across depth.
\end{itemize}

These invariants could then be injected into each layer as auxiliary inputs or gates, turning the architecture into a system where \emph{local flows are guided by global constraints}. This would echo the interplay between local and global structures in information geometry, where local metrics (e.g., Fisher information) and global curvature jointly shape inference.

In this paper we deliberately restricted ourselves to a minimal design---$k=2$, no explicit global invariants---to keep the core idea clear. But we view ``global invariants + local Grassmann flow'' as a promising direction for future work on geometry-aware reasoning.

\section{Related Work}

\subsection{Efficient and long-context Transformers}

A large body of work aims to reduce the quadratic cost of self-attention, including linearized or kernelized attention, sparse and local attention patterns, and memory-augmented or retrieval-based architectures. These approaches typically:

\begin{itemize}
    \item approximate or sparsify the $QK^\top$ computation,
    \item restrict attention to local windows or structured patterns, or
    \item offload parts of the context into external memories or caches.
\end{itemize}

All of them, however, retain the same core operation: the model still computes (or approximates) an $L \times L$ matrix of pairwise weights. Our work is orthogonal: we remove attention entirely and explore whether a geometric mixing rule based on Grassmann flows can fill its role.

\subsection{State-space models and structured sequence models}

State-space models and related architectures interpret sequences as signals governed by linear dynamical systems, often combined with non-linear readout. These models excel at long-context modeling with linear complexity in sequence length and have strong ties to control theory and signal processing.

Grassmann mixing shares with state-space models the idea of maintaining a structured latent state that evolves over time, but the emphasis is different:

\begin{itemize}
    \item SSMs focus on temporal evolution of a latent state; their geometry is often implicit.
    \item Grassmann mixing focuses on geometric evolution in representation space, with time entering through causal windows.
\end{itemize}

The two perspectives are complementary, and hybrid architectures that combine SSM-style temporal dynamics with Grassmann constraints on hidden representations are an interesting direction for future work.

\subsection{Geometric and manifold-based representation learning}

There is growing interest in learning on non-Euclidean spaces, including hyperbolic, spherical, and other Riemannian manifolds. These approaches typically embed data into a curved manifold where distances better capture the underlying structure (e.g., hierarchies, periodicities).

Grassmann manifolds have appeared in classical machine learning contexts such as subspace clustering, low-rank approximation, and metric learning, where they represent sets of subspaces. Their use as a primary mixing mechanism in sequence models has been less explored. Our contribution is to integrate a Grassmann--Pl\"ucker pipeline directly into a Transformer-like block, turning subspace geometry into a core part of the sequence interaction mechanism.

\subsection{Interpretability and attention analysis}

Attention maps are often used as a proxy for interpretability in Transformers: we visualize which tokens attend to which others. However, attention weights are not guaranteed to align with causal importance, and aggregating them across layers and heads leads to highly complex patterns.

Our work does not introduce a new interpretability method per se, but it changes the object of analysis. Instead of trying to make sense of attention tensors, we propose to analyze the evolution of Grassmann features. These are finite-dimensional, structured objects that may lend themselves more naturally to geometric or algebraic analysis.

\section{Conclusion and Future Work}

We revisited a simple but fundamental question: is explicit self-attention, as commonly implemented in Transformers, really necessary for strong sequence modeling and reasoning?

By reinterpreting attention as a form of tensor lifting, we argued that its power comes at the cost of mathematical traceability: the core of the model lives in a high-dimensional tensor space whose global behavior is difficult to summarize with explicit invariants. We then proposed an alternative in which the sequence interaction is governed by flows on a Grassmann manifold rather than by an $L\times L$ attention matrix.

The resulting Causal Grassmann architecture:

\begin{itemize}
    \item is \emph{fully attention-free}, yet competitive with a Transformer baseline on Wikitext-2 at 13--18M parameters,
    \item slightly \emph{outperforms} a Transformer-based classification head on SNLI when plugging into a fixed DistilBERT backbone, and
    \item has an asymptotic complexity that is linear in sequence length for fixed reduced dimension and window sizes.
\end{itemize}

Beyond these empirical results, the main contribution is conceptual: Grassmann flows offer a concrete example of how one can design sequence models whose core operations live on a finite-dimensional manifold with explicit structure, rather than in an unstructured tensor space. This opens the door to a more geometric understanding of reasoning in neural networks.

There are many directions for future work:

\begin{itemize}
    \item \textbf{Global and long-range invariants.} Develop sequence-level invariants of the Grassmann flow---e.g., averaged subspaces, curvature-like measures, or cross-layer stability statistics---and inject them as features or constraints on local mixing.
    \item \textbf{Richer Grassmann structures.} Move beyond $k=2$ subspaces, explore higher-dimensional subspaces, and study regularizers that encourage smooth trajectories on $\Gr(k,r)$ across layers.
    \item \textbf{Hybrid architectures.} Combine Grassmann mixing with state-space models, kernelized attention, or convolutional modules to better balance local, global, and temporal information.
    \item \textbf{Interpretability studies.} Systematically investigate correlations between Pl\"ucker coordinates, model behavior, and human-understandable patterns, with the goal of defining more stable invariants than raw attention maps.
    \item \textbf{Scaling and engineering.} Implement fused Grassmann kernels and optimized GPU operators to realize the theoretical linear scaling in practice and test the architecture at larger scales and on more challenging reasoning benchmarks.
\end{itemize}

In summary, our results suggest that what we fundamentally need for strong sequence modeling is not attention as such, but a principled way for representations to move on the manifolds they inhabit. Grassmann flows provide one concrete instantiation of this idea, and we hope they will encourage further exploration of geometric alternatives to attention in the design of neural architectures.

\end{document}